\documentclass[sigconf]{acmart}

\setcopyright{none}
\renewcommand\footnotetextcopyrightpermission[1]{}
\acmYear{2026}
\acmConference[MiLeTS '26]{The 12th Mining and Learning from Time Series Workshop}{August 2026}{Jeju, Republic of Korea}
\acmBooktitle{The 12th Mining and Learning from Time Series Workshop (MiLeTS '26), held in conjunction with KDD 2026}

\usepackage{mathtools}
\usepackage{array}
\usepackage{tikz}
\usetikzlibrary{arrows.meta, positioning, fit, calc, backgrounds}

\title[Aionoscope Diagnostic Tool]{Aionoscope: Debugging Latent-State Accessibility\\in Time-Series Representations}

\author{Alexander Chemeris}
\affiliation{
  \institution{Langotime}
  \country{South Africa}
}
\email{alex@langotime.ai}

\author{Ming Jin}
\affiliation{
  \institution{Griffith University}
  \country{Australia}
}

\author{Randall Balestriero}
\affiliation{
  \institution{Brown University}
  \country{United States}
}

\begin{document}

\begin{abstract}
Time-series models are often evaluated by what they can forecast or classify, but those scores do not show whether their representations preserve the process state a user may want to inspect: event timing, phase, amplitude, frequency, or regime variables. We introduce Aionoscope, a generator-based diagnostic tool for debugging latent-state accessibility in frozen time-series representations. Aionoscope separates process generation from observation rendering, producing seeded synthetic streams with exact categorical and dense labels across mixture complexity and nuisance variation.

We instantiate Aionoscope as Primitive Process Mixtures and evaluate 37 model-plus-adapter systems with a common pooled linear-probe protocol. The main result is a mismatch between coarse and fine-grained accessibility. Most systems make component presence easy to recover, but expose dense process state much less reliably: the highest observed dense-probe row reaches 0.689 mean masked $R^2$, while a dense-feature oracle reaches 0.999. This is the failure mode Aionoscope is designed to surface: a representation can look informative at the level of ``what kind of signal is present'' while hiding the timing, phase, amplitude, frequency, or regime variables needed for debugging.

This first public validation-seed snapshot is a diagnostic unit test, not a stable leaderboard or a real-task transfer claim. Its rows are conditional on native-length adapters, one train stream, one probe-seed schedule, public validation streams, and the specified pooled linear readout. Detailed interactive results are available at \url{https://aionoscope.langotime.ai/}; source artifacts are available for \href{https://github.com/langotime/aionoscope/}{Aionoscope} and the \href{https://github.com/langotime/aionoscope-benchmarks/}{benchmark repository}.
\end{abstract}



\maketitle

\section{Introduction}

Two time-series windows can support similar short-horizon forecasts while requiring different decisions: one may contain drift, a phase shift, a level change, or a rare event that the other does not. For forecasting, the next observed value may be enough. For diagnosis, event detection, control, and transfer, the useful object is often the latent process state behind the signal: which components are active, which regime or event is present, and which continuous parameters govern the observation.

This gap is especially important for high-volume operational streams. Monitoring, infrastructure, finance, robotics, scientific sensing, and clinical settings contain large amounts of repetitive normal-state behavior, while failures, interventions, regime changes, and other high-value events occupy a small tail of the distribution. Average forecasting or reconstruction losses can therefore reward representations of common background behavior while compressing away the rare or mixed states that matter for diagnosis and decision support.

Real-world benchmarks remain essential, but they rarely expose the latent process state directly. Their labels are often sparse, delayed, entangled with a particular observation regime, or reused across many models. Aionoscope complements such benchmarks with controlled generator-based debugging: each stream is produced from an explicit latent process, and the same process emits exact categorical and continuous state labels. Real-data benchmarks test ecological validity; Aionoscope tests controlled state accessibility under a fixed readout.

Aionoscope targets this gap by separating three questions: which components are separable, which dense state variables are accessible, and where in the layer stack they are exposed. Primitive Process Mixtures is one controlled generator-based instance of that diagnostic test, not a replacement for broad real-data benchmarks or evidence that scores transfer to operational tasks. Here, diagnostic means a synthetic unit test for pooled linear accessibility that generates hypotheses about which state variables and layers to inspect in later task-specific studies. All empirical claims in this paper are conditional on one train stream, one probe-seed schedule, public validation streams, native-length adapters, and post-selected diagnostic displays.

The paper makes three contributions:
\begin{enumerate}
  \item \textbf{Generator-based diagnostic tool.} Aionoscope introduces a Process-to-View contract for debugging pooled linear access to latent process state under controlled generation.
  \item \textbf{State-accessibility probe protocol.} Aionoscope defines a native-length frozen-representation protocol with exact categorical and dense ground truth, deterministic seeded streams, layer-wise pooled linear probes, and validation-seed summaries.
  \item \textbf{Coarse-vs-dense diagnostic finding.} Primitive Process Mixtures audits 37 systems and shows that representations may expose component presence while hiding dense timing, phase, amplitude, frequency, or regime variables.
\end{enumerate}

\begin{figure*}[t]
  \centering
  \includegraphics[width=0.92\textwidth]{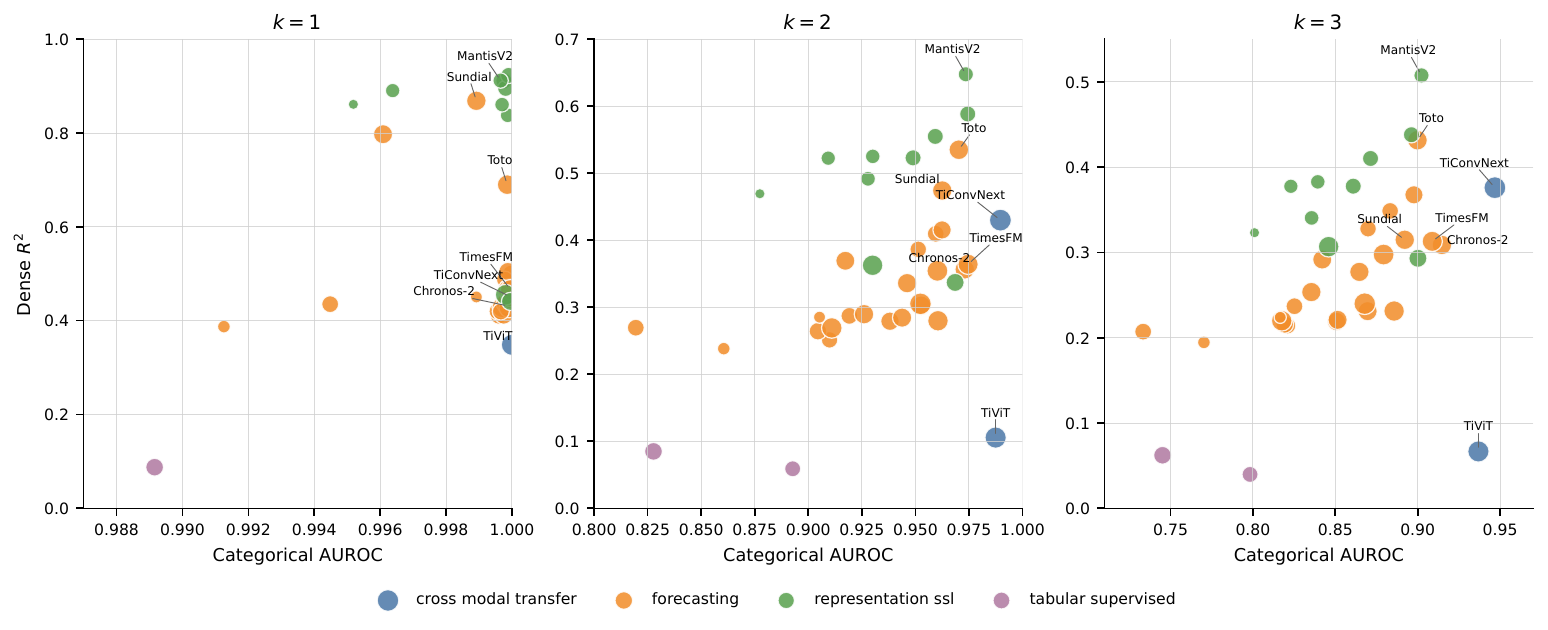}
  \caption{Current-sweep comparison between categorical component AUROC and dense-parameter $R^2$ for $k=1,2,3$. Points summarize one native-length model-plus-adapter system under the common pooled linear-probe readout; point size reflects log parameter count. Panel axes differ because $k=1$ categorical AUROC is near-saturated.}
  \Description{Scatter plot comparing categorical AUROC and dense R-squared across evaluated encoders.}
  \label{fig:scatter-auroc-r2}
\end{figure*}

\section{Benchmark Design}

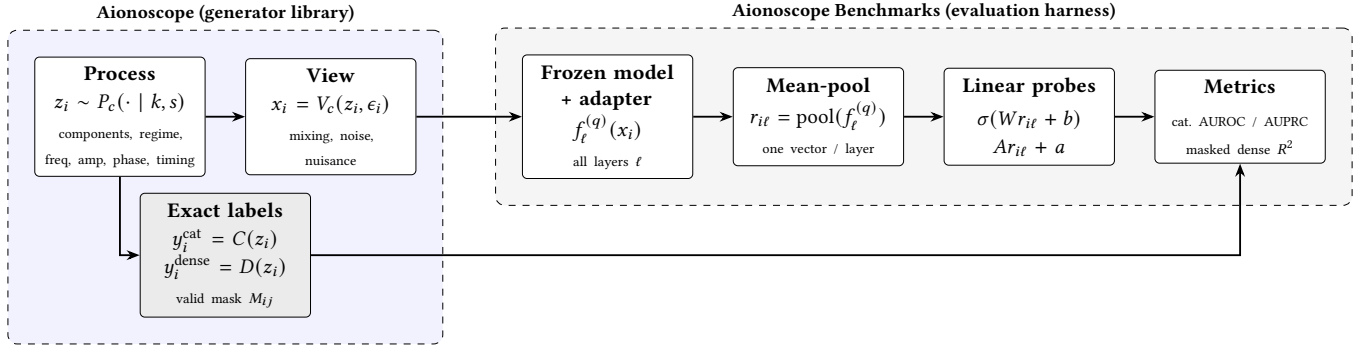
\begin{figure*}[t]
  \centering
  \resizebox{\textwidth}{!}{%
  \begin{tikzpicture}[
    box/.style={draw, rounded corners=2pt, fill=white, align=center, minimum height=12mm, inner sep=4pt, text width=23mm},
    gt/.style={box, fill=black!8},
    meta/.style={draw, dashed, rounded corners=4pt, inner sep=4mm},
    flow/.style={-{Stealth[length=2mm]}, thick},
  ]
    \node[box] (proc) {\textbf{Process}\\[1pt]$z_i\sim P_c(\cdot\mid k,s)$\\[2pt]{\scriptsize components, regime,}\\{\scriptsize freq, amp, phase, timing}};
    \node[box, right=6mm of proc] (view) {\textbf{View}\\[1pt]$x_i=V_c(z_i,\epsilon_i)$\\[2pt]{\scriptsize mixing, noise,}\\{\scriptsize nuisance}};
    \node[gt] (labels) at ($(proc.south)!0.5!(view.south)+(0,-12mm)$) {\textbf{Exact labels}\\[1pt]$y^{\mathrm{cat}}_i=C(z_i)$\\$y^{\mathrm{dense}}_i=D(z_i)$\\[2pt]{\scriptsize valid mask $M_{ij}$}};
    \node[box, right=16mm of view] (model) {\textbf{Frozen model}\\\textbf{+ adapter}\\[1pt]$f^{(q)}_\ell(x_i)$\\[2pt]{\scriptsize all layers $\ell$}};
    \node[box, right=6mm of model] (pool) {\textbf{Mean-pool}\\[1pt]$r_{i\ell}=\mathrm{pool}(f^{(q)}_\ell)$\\[2pt]{\scriptsize one vector / layer}};
    \node[box, right=6mm of pool] (probe) {\textbf{Linear probes}\\[3pt]$\sigma(W r_{i\ell}+b)$\\[1pt]$A r_{i\ell}+a$};
    \node[box, right=6mm of probe] (metric) {\textbf{Metrics}\\[3pt]{\scriptsize cat.\ AUROC / AUPRC}\\[1pt]{\scriptsize masked dense $R^2$}};

    \draw[flow] (proc)--(view);
    \draw[flow] (view)--(model);
    \draw[flow] (model)--(pool);
    \draw[flow] (pool)--(probe);
    \draw[flow] (probe)--(metric);
    \draw[flow] (proc.south) |- (labels.west);
    \draw[flow] (labels.east) -| (metric.south);

    \begin{scope}[on background layer]
      \node[meta, fill=blue!5, fit=(proc)(view)(labels),
        label={[font=\small\bfseries]above:Aionoscope (generator library)}] {};
      \node[meta, fill=black!4, fit=(model)(pool)(probe)(metric),
        label={[font=\small\bfseries]above:Aionoscope Benchmarks (evaluation harness)}] {};
    \end{scope}
  \end{tikzpicture}%
  }
  \caption{End-to-end Aionoscope pipeline under the Process-to-View contract, split into its two released components. The \textbf{generator library} (left) samples a latent state $z_i$ via the \emph{Process}, renders it into an observed stream $x_i$ via the \emph{View}, and emits ground truth from the \emph{same} state ($y^{\mathrm{cat}}_i=C(z_i)$, $y^{\mathrm{dense}}_i=D(z_i)$, with valid mask $M_{ij}$). The \textbf{Benchmarks} harness (right) extracts per-layer features from a frozen model-plus-adapter, mean-pools them into one vector $r_{i\ell}$ per sample and layer, fits categorical and masked-dense linear probes, and scores them against the library's exact labels. Because the observed signal and its labels share one source of truth, accessibility is measured against exact latent state rather than proxy annotations.}
  \Description{Block diagram of the Aionoscope pipeline: a process block emits a latent state, a view block renders an observed series, a frozen model-plus-adapter extracts per-layer features, mean pooling produces one vector per layer, linear probes predict categorical and dense targets, and a metrics block compares predictions against exact labels emitted by the same process.}
  \label{fig:pipeline}
\end{figure*}

Aionoscope follows a Process-to-View factorization. The \emph{Process} defines the latent state: active components, regimes, events, frequencies, amplitudes, slopes, offsets, phase, timing, and other generative parameters. The \emph{View} renders that state into an observed time series through mixing, nuisance variation, noise, and other observation effects. The benchmark then asks whether a frozen representation preserves the Process state rather than only the rendered View. Figure~\ref{fig:pipeline} summarizes the full pipeline, from process sampling to probe scoring against exact labels.

\paragraph{Formal protocol.}
For benchmark configuration $c$, seed $s$, and mixture complexity $k$, Aionoscope samples latent process states and renders views
\[
z_i \sim P_c(\cdot \mid k,s),\quad x_i = V_c(z_i,\epsilon_i),\quad
y_i^{\mathrm{cat}} = C(z_i),\quad y_i^{\mathrm{dense}} = D(z_i).
\]
Here $y_i^{\mathrm{cat}}\in\{0,1\}^{14}$ marks active components and $y_i^{\mathrm{dense}}\in\mathbb{R}^{34}\cup\{\mathrm{NaN}\}$ stores dense parameters. A mask $M_{ij}=1$ only when target $j$'s owning component is active. For model $q$ and layer $\ell$, the adapter extracts $r_{i\ell}^{(q)}=\mathrm{pool}(f_\ell^{(q)}(x_i))$ and fits linear probes
\[
\hat y_{i\ell}^{\mathrm{cat}}=\sigma(W_\ell r_{i\ell}^{(q)}+b_\ell),\qquad
\hat y_{i\ell}^{\mathrm{dense}}=A_\ell r_{i\ell}^{(q)}+a_\ell,
\]
with dense losses and metrics evaluated only where $M_{ij}=1$.

\begin{center}
\begin{scriptsize}
\begin{tabular}{@{}rl@{}}
1 & for each model $q$ and $k\in\{1,2,3\}$ resolve the native input length; \\
2 & generate train stream with seed 0 and validation streams with seeds 100--109; \\
3 & extract frozen representations from every available layer; \\
4 & pool each layer representation into one vector per sample; \\
5 & fit categorical and masked dense linear probes with probe seed 0; \\
6 & score each layer, target group, and validation stream; \\
7 & report validation-seed medians per run and mean summaries across $k$. \\
\end{tabular}
\end{scriptsize}
\end{center}

The first benchmark instance, Primitive Process Mixtures, evaluates single-channel synthetic streams built from 14 component processes: a constant baseline; three noise components (Gaussian noise, uniform noise, and random-walk noise); four trends (linear, quadratic, log, and sigmoid); three periodic processes (sine, sawtooth, and square); and three localized events (spike, level change, and Gaussian pulse). Its technical tag is \texttt{aiono\_basic\_components}\slash\texttt{v2}. Each generated sample carries two target types: 14 categorical component-state labels and 34 dense generative parameters covering timing, magnitude, noise scale, trend coefficients, frequency, phase, offset, and duty cycle. We evaluate mixture complexity with $k \in \{1,2,3\}$ active components per sample. Under the balanced component sampler, each component has marginal prevalence $k/14$: 7.1\%, 14.3\%, and 21.4\% for $k=1,2,3$ before finite-stream variation.

Mixture complexity is part of the benchmark rather than an incidental nuisance. At $k=1$, many models can identify isolated primitive patterns. At $k=2$ and $k=3$, multiple processes interfere in the same observed signal, which better matches event and regime diagnosis where useful states are embedded in background variation. This makes Aionoscope a controlled comparison setting for pattern detection, anomaly and event diagnosis, representation comparison, and layer-wise interpretability.

Each benchmark stream is deterministic given the benchmark version, generator configuration, and seed, enabling exact regeneration under the recorded execution stack and construction of future held-out streams. The current validation seeds were fixed before the artifact release and were not available as public targets before this submission; after publication they become public reference/development seeds, not a locked test set. Future public benchmark claims should add hidden or locked evaluation streams. The implementation materializes streams and aligned labels from the same seed-controlled generator, avoiding dependence on a fixed stored corpus and keeping labels attached to the generating process, without making runtime or throughput claims in this paper.

\begin{table*}[t]
\caption{Primitive Process Mixtures contract used for the reported sweep. This is the deterministic benchmark contract for the current paper, not a claim that the benchmark family is complete.}
\label{tab:benchmark-contract}
\centering
\begin{scriptsize}
\begin{tabular}{lp{0.76\textwidth}}
\toprule
Field & Value \\
\midrule
Benchmark instance & \texttt{aiono\_basic\_components/v2}, single-channel \texttt{mix} view, 500 Hz sampling. \\
Components & 14 labels: constant; Gaussian, uniform, and random-walk noise; linear, quadratic, log, and sigmoid trends; sine, sawtooth, and square periodic components; spike, level-change, and Gaussian-pulse events. \\
Mixture settings & $k \in \{1,2,3\}$ active components per sample; balanced marginal prevalence per component is $k/14$. \\
Sequence length & Each adapter's model-native exact length; current sweep spans 128--16,384 samples, or 0.256--32.768 s at 500 Hz. \\
Train/validation streams & Train seed 0 with $256 \times 256=65{,}536$ samples. Validation seed values 0--9 map to generator seeds 100--109, with 65,536 samples per validation seed. \\
Probe settings & Linear probe, 1,250 AdamW steps, batch size 1,024, learning rate 0.01, no weight decay, gradient clip 1.0, probe seed 0. \\
Aggregation & Run-level values are medians across the 10 validation seeds; mean columns average the three run-level medians for $k=1,2,3$. \\
\bottomrule
\end{tabular}
\end{scriptsize}
\end{table*}

\paragraph{Dense target schema.}
The 34 dense targets are generated by the same process/view objects that render the signal. Event times are normalized sample indices in $[0,1]$ and are sampled from the central 15--85\% of the sequence. Event magnitudes use spike amplitude in $[0.8,1.2]$, level-change and Gaussian-pulse amplitudes in $[-1,1]$, and Gaussian-pulse width in $[0.01,0.06]$ seconds. Noise targets are Gaussian standard deviation in $[0.02,0.15]$, uniform-noise amplitude in $[0.05,0.25]$, and random-walk step standard deviation in $[0.01,0.08]$. Trend targets cover linear slope/intercept in $[-2,2]$ and $[-0.5,0.5]$, quadratic $a,b,c$ in $[-4,4]$, $[-2,2]$, and $[-0.5,0.5]$, log-trend amplitude/offset in $[-2,2]$ and $[-0.5,0.5]$, and sigmoid amplitude/center/sharpness/offset in $[-2,2]$, $[0.2,0.8]$, $[5,20]$, and $[-0.5,0.5]$. Periodic amplitude, phase, and offset use $[0.2,1.2]$, $[0,2\pi]$, and $[-0.2,0.2]$; square duty cycle uses $[0.1,0.9]$. Periodic frequency bounds are resolved per model-native sequence length: the lower bound gives at least one full period, sine uses 0.9 Nyquist, sawtooth requires at least five points per period, and square waves require at least two points in the shorter duty-cycle plateau. These targets are resolved after native length is known, so event time fractions remain comparable while absolute clock-time windows change at 500 Hz, and periodic rows face related but not identical frequency distributions.

\section{Evaluation Protocol}

For each model, the adapter resolves the model-native input length, generates train and validation streams at that length, extracts frozen features from adapter-exposed layers, and fits lightweight probes. The current protocol uses linear probes over mean-pooled layer representations. The primary descriptive endpoints are mean categorical AUROC and mean dense $R^2$, each averaged over $k=1,2,3$; AUPRC, Pearson correlation, $k=3$ results, and layer curves are secondary diagnostics. Categorical AUROC and AUPRC are macro-averaged over the 14 component labels. Dense targets are finite only when the owning component is active; inactive-component values are masked as missing and excluded from regression loss and metrics. For dense target $j$, validation $R^2_j=1-\sum_{i:M_{ij}=1}(y_{ij}-\hat y_{ij})^2/\sum_{i:M_{ij}=1}(y_{ij}-\bar y_j)^2$ on the masked validation subset, so negative values mean the probe is worse than predicting that subset's target mean. Macro dense $R^2$ and Pearson average the 34 per-target metrics after this masking, so they should be read as equal-weight diagnostic summaries rather than unit-aware physical errors.

A row in the current sweep is a model-plus-adapter fingerprint: it includes the model, adapter extraction path, preprocessing, native input length, and domain match. We intentionally use each adapter's model-native exact input length to test released systems in the most documented and adapter-faithful regime available, while avoiding extra padding, truncation, resampling, or windowing artifacts. This is the intended diagnostic setting for the released systems, but it makes native length part of what is evaluated. At the fixed 500 Hz view, native lengths imply clock-time windows from 0.256 to 32.768 seconds, and many forecasting checkpoints were pretrained mainly on slower calendar-time domains. The sweep is therefore a heterogeneous model-plus-adapter-plus-domain-match audit, not a length-controlled architecture comparison. As descriptive checks on this same snapshot, log native length has only weak association with mean dense $R^2$ across the 37 rows (Spearman 0.08; Pearson 0.10), and collapsing to one highest-dense row per coarse model family leaves AUROC--dense $R^2$ Spearman at 0.31 over 21 families. These checks do not remove length, family, or adapter confounding; they only show that the reported patterns are not reducible to a simple monotone length story. Clustering of related systems is hypothesis-generating, not causal evidence that an architecture or objective is better. Validation-seed summaries reuse one fixed train stream and one deterministic probe-seed schedule; sweeping train or probe seeds would multiply cost, so we leave that robustness audit to future work. The Spearman interval uses a Fisher transform over the 37 system rows only and does not include train-stream, probe-seed, checkpoint, adapter, model-family dependence, or benchmark-spec uncertainty. We treat mean summaries as primary diagnostics, while family interpretations, per-target upper envelopes, and selected layer/radar views are exploratory.

To calibrate task difficulty without mixing hand-built rows into the pretrained-system inventory, we also run fixed-feature baselines at controlled lengths $L\in\{128,512,2048,5000,16384\}$. These include a metric floor, raw downsampled waveform features, a fixed random projection of the waveform, FFT log-power features, windowed statistics, statistics plus FFT features, and two oracle rows that expose generator labels or dense label features. Baseline rows use synthetic layer 0 and the same train seed, validation seeds, mixture settings, and linear-probe protocol where applicable; they answer how visible the targets are to simple waveform features, not whether a pretrained encoder has learned to expose them. Dense probes fit raw target values without per-target standardization. The copy oracle receives component indicators, dense values, and dense-valid masks as features, and reaches near-perfect $R^2$ through the same linear masked dense-probe path; it is a sanity check, not a waveform baseline.

Mean pooling is a deliberate common denominator. The evaluated adapters expose different internal objects, including token or patch states, CLS summaries, image-rendered features, forecasting-token pools, recurrent states, and tabular probability embeddings. A single pooled vector avoids model-specific heads or privileged readout logic while allowing the same probe protocol to be applied at every adapter-exposed layer.

This protocol is intentionally low-capacity and model-agnostic. If a linear probe succeeds, the state variable is readily accessible from the frozen representation under a simple common readout. If it fails, the conclusion is narrower: the state was not linearly recoverable from this pooled representation. Mean pooling can wash out token-local information, phase or location cues, and information stored in special tokens or model-specific forecast heads. We evaluate model-plus-adapter systems as exposed by the benchmark adapters; adapter redesign or ablation is separate model-author work. CLS-only, mean-only, token-level, learned-pooling, nonlinear, manifold-learning, model-specific-head, or downstream fine-tuning readouts may recover additional information and are natural extensions of the benchmark.

\section{Results}

\paragraph{Interactive results browser.}
The page-limited PDF includes the full metric inventory and selected diagnostics; more detailed model, target, radar, and layer views are available at \url{https://aionoscope.langotime.ai/}.

\paragraph{Source artifacts.}
The source artifacts are available in public repositories: the \href{https://github.com/langotime/aionoscope/}{generator/library artifact} and the \href{https://github.com/langotime/aionoscope-benchmarks/}{benchmark artifact}. The benchmark artifact contains the model/checkpoint registry, adapter paths, configs, scripts, manifests, result schema, frozen JSON results, and dashboard assets; the browser is a viewer over that snapshot rather than the sole source of results.

\paragraph{Interpretation guardrails.}
The results below describe heterogeneous model-plus-adapter systems at native lengths, not controlled encoder-quality estimates. Success means pooled linear accessibility; failure means only no recovery by this readout. The metric inventory is display-oriented; selected layer/radar views, target maxima, and family interpretations are exploratory. Validation-seed intervals are not rank-stability or pairwise-comparison intervals.

\paragraph{Calibration baselines.}
Table~\ref{tab:baseline-calibration} reports fixed-feature calibration rows separately from the pretrained-system inventory, including the completed long settings at $L=5000$ and $L=16384$ plus ranges over all five controlled lengths. Raw waveform and random-projection rows stay near categorical floor. FFT log-power and statistics-plus-FFT features give strong categorical anchors but poor dense macro recovery under the common raw-target regression probe, especially at long lengths. Windowed statistics are the strongest non-oracle dense classical anchor, reaching 0.274--0.525 mean dense $R^2$ across controlled lengths. The oracle rows sanity-check different probe-stack parts: component indicators solve categorical labels but not dense parameters, while label features reach 0.999 mean dense $R^2$. Thus the dense probe can recover the target channels when they are directly exposed, and model-plus-adapter dense failures should not be explained by an inability of the probe stack to copy dense state.
Large negative dense $R^2$ values mean that a raw-target shared probe can perform worse than the target-mean baseline on masked subsets; they are a calibration warning, not a generator failure. These fixed-feature rows also do not provide a supervised waveform upper bound.

\begin{table*}[t]
\caption{Fixed-feature calibration baselines. Metrics average the three $k=1,2,3$ run medians. The $L=5000$ and $L=16384$ columns show two completed long-length calibration settings; range columns show min--max over $L\in\{128,512,2048,5000,16384\}$. These rows calibrate task difficulty and sanity-check the common probe stack; they are not included in the pretrained-system metric inventory. The dense-feature oracle exposes component indicators, active dense parameter values with inactive entries zero-filled, and dense-valid masks, and verifies that the same linear masked dense-probe path can recover exposed dense state nearly perfectly.}
\label{tab:baseline-calibration}
\centering
\begin{tiny}
\setlength{\tabcolsep}{2.7pt}
\begin{tabular}{llrrrrrr}
\toprule
Row & Role & \multicolumn{2}{c}{$L=5000$} & \multicolumn{2}{c}{$L=16384$} & \multicolumn{2}{c}{All-length range} \\
 & & AUROC & $R^2$ & AUROC & $R^2$ & AUROC & $R^2$ \\
\midrule
MetricFloor & prevalence / mean target & 0.500 & 0.000 & 0.500 & 0.000 & [0.500, 0.500] & [0.000, 0.000] \\
RawDownsample512 & raw waveform & 0.501 & $-41.480$ & 0.501 & $-142.775$ & [0.501, 0.503] & [$-142.775$, 0.100] \\
RandomProjection512 & random waveform projection & 0.502 & $-1.108$ & 0.501 & $-1.497$ & [0.501, 0.505] & [$-1.497$, $-0.387$] \\
FFTLogPower512 & spectral log-power & 0.836 & $-6.497$ & 0.801 & $-219.712$ & [0.801, 0.858] & [$-219.712$, 0.121] \\
StatsWindowed & time-domain statistics & 0.874 & 0.455 & 0.869 & 0.274 & [0.842, 0.874] & [0.274, 0.525] \\
StatsFFTCombined & statistics + spectral features & 0.932 & $-21.556$ & 0.926 & $-66.593$ & [0.897, 0.932] & [$-66.593$, $-7.006$] \\
OracleEnabledMask & component-mask oracle & 1.000 & $-0.051$ & 1.000 & $-0.051$ & [1.000, 1.000] & [$-0.058$, $-0.051$] \\
OracleDenseParams & dense-feature oracle & 1.000 & 0.999 & 1.000 & 0.999 & [1.000, 1.000] & [0.999, 0.999] \\
\bottomrule
\end{tabular}
\end{tiny}
\end{table*}

\begin{table*}[t]
\caption{Representative model results. Mean columns average the three median scores for $k=1,2,3$ enabled components. The $k=3$ columns report the hardest mixture setting.}
\label{tab:key-results}
\centering
\begin{scriptsize}
\begin{tabular}{lllrrrrr}
\toprule
Model & Backbone & Training & Params & Mean AUROC & Mean $R^2$ & AUROC $k=3$ & $R^2$ $k=3$ \\
\midrule
TiConvNext & vision convnet & cross modal transfer & 1.20B & 0.979 & 0.419 & 0.947 & 0.376 \\
TiViT & vision transformer & cross modal transfer & 630.8M & 0.975 & 0.174 & 0.937 & 0.067 \\
Chronos-2 & transformer full attention & forecasting & 119.5M & 0.963 & 0.365 & 0.915 & 0.309 \\
MantisV2 & transformer full attention & representation ssl & 4.2M & 0.959 & 0.689 & 0.902 & 0.508 \\
TimesFM & transformer causal & forecasting & 231.3M & 0.961 & 0.381 & 0.909 & 0.313 \\
MantisPlus & transformer full attention & representation ssl & 8.1M & 0.957 & 0.650 & 0.896 & 0.438 \\
Toto & transformer full attention & forecasting & 151.3M & 0.957 & 0.552 & 0.900 & 0.432 \\
Mantis-UTICA & transformer full attention & representation ssl & 8.1M & 0.937 & 0.599 & 0.861 & 0.378 \\
Mantis-8M & transformer full attention & representation ssl & 8.1M & 0.943 & 0.625 & 0.871 & 0.410 \\
NuTime-Bias9 & transformer full attention & representation ssl & 2.4M & 0.910 & 0.597 & 0.823 & 0.377 \\
UniShape-ZeroShot & transformer full attention & representation ssl & 2.8M & 0.923 & 0.589 & 0.839 & 0.383 \\
UniShape-FineTune & transformer full attention & representation ssl & 3.1M & 0.921 & 0.557 & 0.836 & 0.340 \\
\bottomrule
\end{tabular}
\end{scriptsize}
\end{table*}

\begin{figure*}[t]
  \centering
  \includegraphics[width=0.92\textwidth]{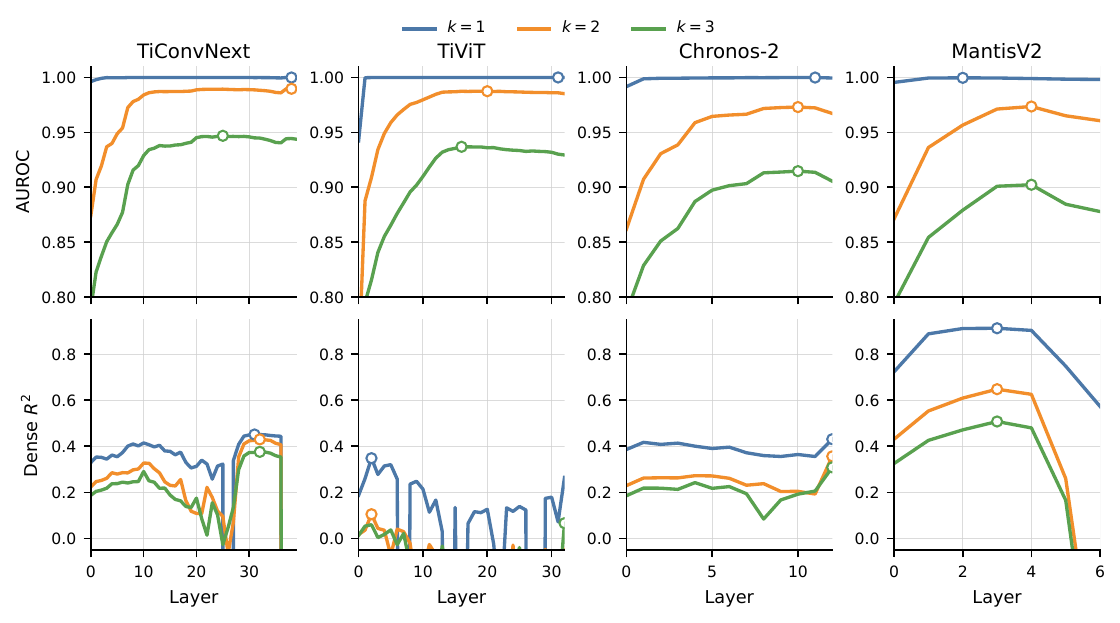}
  \caption{Layer-wise pooled linear probes for four contrastive systems selected after inspection to span high-categorical image-transfer, high-categorical forecasting, and high-dense regions of the diagnostic plane. Open circles mark each curve maximum; dense $R^2$ panels clip values below $-0.05$ only for readability, so clipped layers indicate severe readout failure rather than equal performance.}
  \Description{Line plots of layer-wise categorical and dense probe performance for selected models and mixture complexities.}
  \label{fig:layer-curves}
\end{figure*}

\begin{figure*}[t]
  \centering
  \includegraphics[width=0.96\textwidth]{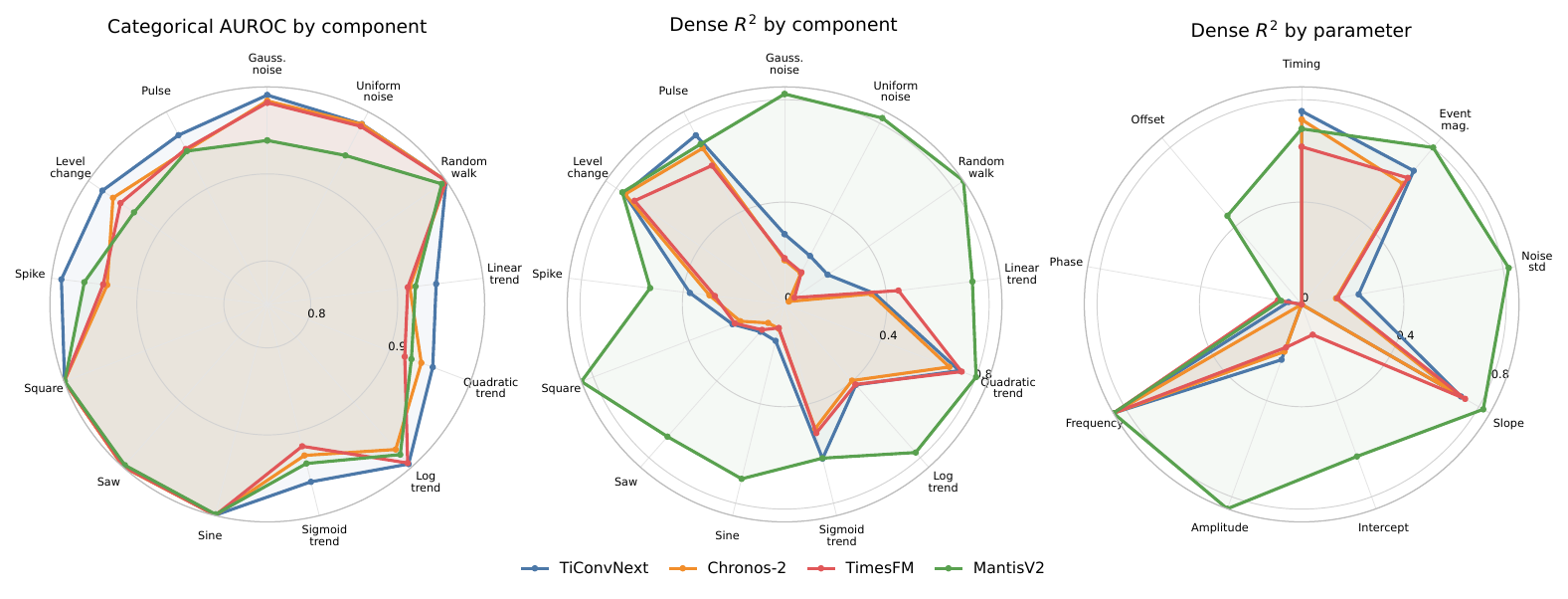}
  \caption{Post-selection diagnostic fingerprints for the same contrastive systems used in the layer figure; these are illustrative slices, not a random or representative sample of the 37-system sweep. Values average the $k=1,2,3$ run medians. The categorical panel uses each run's best-AUROC layer; dense panels use each run's best macro-$R^2$ layer and clip negative $R^2$ to zero for display. These radars are debugging views of where a representation exposes state under the pooled linear readout, not a separate leaderboard; full per-model radars are available in the interactive results browser.}
  \Description{Radar plots comparing categorical and dense target-family fingerprints for selected encoders.}
  \label{fig:diagnostic-radar}
\end{figure*}

\paragraph{Exploratory observations.}
The following observations summarize one public validation-seed snapshot. They were not preregistered replication tests, and the layer/radar figures were selected after inspection to illustrate contrastive regions of the diagnostic plane. Their role is to identify failure modes and follow-up hypotheses, not to establish stable model rankings.

\paragraph{Observation 1: categorical and dense summaries are not interchangeable.}
The same system-plus-adapter audit can place rows in different parts of the categorical and dense-probe plane. Image-transfer rows occupy a high-categorical region in this snapshot, while several related Mantis-family rows occupy a high-dense region under the same pooled readout. Across all 37 systems, the exploratory Spearman association between mean categorical AUROC and mean dense $R^2$ is 0.422, with an approximate Fisher-transform 95\% interval of 0.114--0.657. When collapsed to one highest-dense row per coarse model family, the association remains positive but weaker at 0.31 over 21 families. Because length, adapter, checkpoint, model-family dependence, and seed uncertainty are not controlled, these numbers motivate reporting both endpoints; they are not stable adjacent-rank evidence or a single model winner.

\paragraph{Observation 2: target fingerprints expose different failure profiles.}
Figure~\ref{fig:diagnostic-radar} shows why the sweep is better read as a debugging tool than as a single leaderboard. The categorical panel is high for several systems on many visible components, but the dense panels separate parameter families: MantisV2 has a broader dense footprint on noise, trend, and periodic parameters, while TiConvNext's categorical strength does not translate uniformly to dense scale, offset, phase, and event-parameter recovery. These per-family fingerprints help identify what a model-plus-adapter system recovers, fails to recover, or exposes only weakly under the common pooled linear readout.

\paragraph{Observation 3: waveform-to-image systems form a high-categorical cluster.}
Waveform-to-image systems transferred into the benchmark at their native 5000-sample length form a high-categorical cluster in this snapshot. This is not evidence that image-native inductive bias is generally better for time series: the observation is confounded with model scale, pretraining data, waveform-to-image preprocessing, native length, and the 500 Hz synthetic view. The clustering is a diagnostic hypothesis for length- and adapter-controlled follow-up, not an architecture attribution.

\paragraph{Observation 4: dense-high rows are related.}
In this sweep, the high masked dense-probe region contains several related time-series representation systems, including Mantis-family checkpoints. This should be read primarily as a family-level observation rather than independent evidence that SSL alone causes dense-state retention. Architecture, objective, pretraining data, adapter extraction, and native length co-vary here, so this is a hypothesis about representation-system lineages rather than a causal result. Related checkpoints with different training mixtures also move noticeably in the dense ordering, making dataset choice itself an important diagnostic variable.

\paragraph{Observation 5: layer-wise accessibility varies beyond the selected models.}
The selected layer curves are not monotone. This is not only a visual property of the four plotted models: across the 111 model--$k$ runs, the best dense-$R^2$ layer is non-final in 88 runs and in at least one run for 33 of 37 models; among those non-final dense cases, the median best-to-final drop is 0.050 $R^2$ and 84 of 88 drops exceed 0.005. For AUROC, non-final best layers are also common but the median drop is only 0.003, so the categorical layer effect is often smaller. The plotted curves are selected diagnostic slices, with full per-system layer views delegated to the interactive browser. This pattern matters because many downstream evaluations read only a final representation.

\paragraph{Observation 6: mixture complexity reveals failures hidden at $k=1$.}
With one active component, categorical recovery is close to saturated for many models. Even after averaging over $k=1,2,3$, 32 of 37 systems exceed 0.900 mean AUROC, 20 exceed 0.930, and 10 exceed 0.950, so adjacent categorical rows should not be overread. The more diagnostic regime is $k=3$, where multiple processes interfere in the same observation. At this hardest setting, high-categorical rows remain near 0.94 AUROC, but dense recovery and layer behavior continue to separate systems.

\begin{table*}[t]
\caption{Dense target recovery by generative parameter at $k=3$. The max value is an uncorrected post-selection upper envelope across systems and layers selected on the same validation snapshot; system names are omitted to avoid implying stable winners. The final column reports the median and interquartile range across each system's best layer for that target. This is an exploratory diagnostic of whether any evaluated system made a target linearly accessible, not typical per-model performance or a multiple-comparison-adjusted estimate.}
\label{tab:dense-targets}
\centering
\begin{tiny}
\setlength{\tabcolsep}{3.4pt}
\begin{tabular}{lllrl}
\toprule
Target & Signal & Parameter & Max $R^2$ & System med. [IQR] \\
\midrule
log\_trend\_amplitude & log\_trend & amplitude & 0.983 & 0.891 [0.873,0.921] \\
sine\_frequency\_hz & sine & frequency\_hz & 0.968 & 0.907 [0.844,0.945] \\
square\_frequency\_hz & square & frequency\_hz & 0.966 & 0.894 [0.826,0.923] \\
square\_amplitude & square & amplitude & 0.959 & 0.333 [0.295,0.767] \\
sawtooth\_frequency\_hz & sawtooth & frequency\_hz & 0.947 & 0.825 [0.770,0.896] \\
sine\_amplitude & sine & amplitude & 0.938 & 0.349 [0.290,0.785] \\
square\_duty\_cycle & square & duty\_cycle & 0.936 & 0.860 [0.788,0.899] \\
sawtooth\_amplitude & sawtooth & amplitude & 0.930 & 0.344 [0.292,0.714] \\
random\_walk\_step\_std & random\_walk\_noise & std & 0.868 & 0.184 [0.034,0.352] \\
quadratic\_trend\_a & quadratic\_trend & a & 0.796 & 0.700 [0.666,0.742] \\
sigmoid\_trend\_amplitude & sigmoid\_trend & amplitude & 0.772 & 0.640 [0.603,0.677] \\
gaussian\_noise\_std & gaussian\_noise & std & 0.752 & 0.203 [0.122,0.362] \\
level\_change\_amplitude & level\_change & magnitude & 0.720 & 0.579 [0.520,0.654] \\
uniform\_noise\_amplitude & uniform\_noise & amplitude & 0.706 & 0.203 [0.105,0.363] \\
linear\_trend\_slope & linear\_trend & slope & 0.687 & 0.629 [0.596,0.660] \\
level\_change\_time\_frac & level\_change & time\_frac & 0.674 & 0.297 [0.207,0.405] \\
gaussian\_amplitude & Gaussian pulse & magnitude & 0.667 & 0.453 [0.319,0.537] \\
quadratic\_trend\_b & quadratic\_trend & slope & 0.645 & 0.590 [0.543,0.617] \\
spike\_time\_frac & spike & time\_frac & 0.553 & 0.161 [0.038,0.259] \\
square\_phase & square & phase & 0.542 & 0.115 [0.038,0.196] \\
gaussian\_time\_frac & Gaussian pulse & time\_frac & 0.532 & 0.236 [0.101,0.314] \\
sawtooth\_phase & sawtooth & phase & 0.517 & 0.072 [0.015,0.197] \\
sigmoid\_trend\_center & sigmoid\_trend & center & 0.414 & 0.250 [0.200,0.289] \\
gaussian\_sigma\_sec & Gaussian pulse & sigma\_sec & 0.371 & -0.028 [-0.090,-0.001] \\
sigmoid\_trend\_sharpness & sigmoid\_trend & sharpness & 0.358 & 0.070 [0.039,0.102] \\
linear\_trend\_intercept & linear\_trend & intercept & 0.326 & -0.001 [-0.011,0.254] \\
quadratic\_trend\_c & quadratic\_trend & intercept & 0.302 & -0.001 [-0.008,0.230] \\
sine\_phase & sine & phase & 0.291 & 0.023 [0.000,0.099] \\
sigmoid\_trend\_offset & sigmoid\_trend & offset & 0.275 & -0.003 [-0.014,0.192] \\
log\_trend\_offset & log\_trend & offset & 0.253 & -0.002 [-0.007,0.169] \\
spike\_amplitude & spike & magnitude & 0.105 & -0.014 [-0.050,-0.002] \\
sawtooth\_offset & sawtooth & offset & 0.090 & -0.002 [-0.030,-0.000] \\
sine\_offset & sine & offset & 0.087 & -0.003 [-0.029,-0.000] \\
square\_offset & square & offset & 0.049 & -0.003 [-0.019,-0.000] \\
\bottomrule
\end{tabular}
\end{tiny}
\end{table*}

\paragraph{Target-level behavior.}
Table~\ref{tab:dense-targets} is an uncorrected post-selection upper envelope over evaluated systems and layers at $k=3$: it asks whether any evaluated system made a target linearly accessible under this protocol, without assigning stable per-target winners. It also reports the median/IQR across systems, because a max-only table would overstate typical recovery. Frequency, amplitude, and duty-cycle targets are often highly recoverable, while offsets, spike amplitude, Gaussian-pulse width, and several timing or phase targets are much harder. This supports using masked dense-probe $R^2$ as a compact diagnostic summary, but it also cautions against interpreting a single macro score as uniform latent-state accessibility. Some dense targets are also intrinsically easier to identify from the rendered view than others: frequency and amplitude leave strong global signatures, while offsets, phases, widths, and event timing can become ambiguous under mixtures or mean-pooled readouts. Low $R^2$ therefore reflects the combined difficulty of the generator target, view identifiability, adapter, representation, and pooled linear probe.

\section{Related Work and Positioning}

Aionoscope is a complement to real-world time-series benchmarks, not a replacement for them. Real datasets test whether models transfer to naturally occurring domains, including clinical and multimodal settings \cite{wagner2020ptbxl,strodthoff2023ptbxlplus,williams2024contextiskey,oh2023ecgqa}. Forecasting-focused foundation-model benchmarks such as GIFT-Eval emphasize broad zero-shot forecast comparison across real datasets \cite{aksu2024gifteval}, while recent benchmarking critiques stress leakage and governance risks in time-series foundation-model evaluation \cite{meyer2025rethinking}. Aionoscope addresses a different question: whether representations preserve known latent process state under controlled ground truth.

Recent work has also begun to audit time-series foundation models as frozen feature extractors and to test how forecasting models transfer across domains or spectral regimes \cite{auer2025forecastingfeatures,karaouli2025foundational,wang2025frequencymatters}. Those studies evaluate real-task transfer or specific shift mechanisms. Aionoscope is complementary: it does not validate downstream deployment, but supplies exact latent-state labels and controlled model-plus-adapter diagnostic slices that real datasets usually lack.

Synthetic time-series generation is also not new. Existing systems use generated series for augmentation or pretraining: TimesFM uses a large time-series corpus, Chronos supplements public data with Gaussian-process synthetic series, and CauKer studies classifiers pretrained from synthetic time series \cite{das2024timesfm,ansari2024chronos,xie2025cauker}. Aionoscope uses generation for a different purpose: not to improve pretraining data, but to create a controlled debugging instrument where the sampled latent state, rendered observation, categorical labels, and dense parameters share one source of truth. Its novelty is the concrete seeded multi-target diagnostic harness, not generation, probing, or latent-factor diagnosis as isolated ideas.

Aionoscope's readout protocol is a probing diagnostic: frozen representations are tested with low-capacity readouts to ask what information is linearly accessible at particular layers \cite{alain2016linearprobes,hewitt2019probes,belinkov2022probing}. The probing literature also explains why these results need careful interpretation: probe success is correlational, probe failure is readout-limited, and probe capacity and controls matter. Aionoscope therefore does not claim that probing is new. Its contribution is to pair a time-series Process-to-View generator with exact categorical and dense latent-state targets, regenerated public validation streams, model-plus-adapter manifests, and a common frozen-probe audit across existing encoders.

\begin{table*}[t]
\caption{Positioning of Aionoscope relative to adjacent evaluation and representation-learning work. ``Fresh'' means seeded generation of new evaluation streams rather than a fixed stored test corpus.}
\label{tab:positioning}
\centering
\begin{scriptsize}
\setlength{\tabcolsep}{2.8pt}
\begin{tabular}{p{0.22\textwidth}ccccc p{0.30\textwidth}}
\toprule
Work/category & Data & Cat. labels & Dense params & Fresh & Frozen probe & Main evaluation question \\
\midrule
Real-data TS benchmarks & real & task-specific & no & no & varies & Transfer to naturally occurring forecasting, classification, or diagnosis tasks. \\
TS SSL methods \cite{yue2022ts2vec,eldele2021tstcc,woo2022cost,tonekaboni2021tnc,li2021dts} & real/synthetic & varies & no & no & sometimes & Learn representations that improve downstream time-series tasks. \\
Synthetic training data \cite{fu2024infoboost,das2024timesfm,ansari2024chronos,xie2025cauker} & generated & optional & optional & yes & no & Improve training data or representation learning with generated series. \\
Probing diagnostics \cite{alain2016linearprobes,hewitt2019probes,belinkov2022probing} & varies & varies & varies & no & yes & Test property recovery from frozen representations. \\
Forecasting FM benchmarks \cite{aksu2024gifteval} & real & no & no & no & no & Compare zero-shot or transfer forecasting accuracy across datasets. \\
Time-series QA/reasoning \cite{oh2023ecgqa} & real/synthetic & answer labels & no & varies & no & Evaluate end-to-end answers from a full encoder/readout/reasoner system. \\
Aionoscope & generated & yes & yes & yes & yes & Test whether frozen representations linearly expose controlled latent state. \\
\bottomrule
\end{tabular}
\end{scriptsize}
\end{table*}

\paragraph{What is new.} Aionoscope's novelty is not generation, probing, or latent-factor analysis in isolation, but their specific combination. Unlike synthetic-data work, which generates series to \emph{train} models, and unlike prior probing benchmarks, whose generative factors are usually only partially known, Aionoscope emits each signal together with its exact categorical \emph{and} dense labels from one source of truth, regenerates fresh seeded evaluation streams instead of reusing a fixed corpus, and applies a single layer-wise frozen-probe audit uniformly across models. to the best of our knowledge, Aionoscope is the only entry combining dense latent-state parameters, fresh seeded streams, and a uniform cross-encoder frozen probe (Table~\ref{tab:positioning}).

Aionoscope also differs from representation-learning methods. TS2Vec, TSTCC, CoST, TNC, and DTS show the value of learned or interpretable time-series representations across downstream tasks \cite{yue2022ts2vec,eldele2021tstcc,woo2022cost,tonekaboni2021tnc,li2021dts}. Aionoscope does not propose a new pretraining objective or claim to be a validated pretraining corpus. Its contribution is a diagnostic generator whose sampled latent state, rendered signal, categorical targets, and dense parameters all come from the same source of truth.

In probing terms, a linear readout over frozen features tests restricted accessibility; it does not prove absence of information or causal use by downstream heads. Finally, Aionoscope sits below end-to-end reasoning or question-answering evaluations by asking whether the time-series representation already contains the state variables that later readouts or reasoning systems would need. This representation-level diagnosis helps separate encoder failures from downstream readout or reasoning failures.

\section{Limitations and Intended Use}

Primitive Process Mixtures is intentionally controlled and single-channel, with a compact set of primitive processes rendered at 500 Hz. Many evaluated checkpoints were pretrained on much slower or domain-specific series, so the sweep includes distribution and sampling-rate mismatch as well as representation behavior. The current instance does not cover the full diversity of domain-specific dynamics, multichannel sensing, irregular sampling, missingness, long-horizon temporal dependencies, or real operational distributions. Success may partly reflect shortcut detection on this synthetic grammar rather than latent-state accessibility in a real deployment setting.

Use Aionoscope for diagnostic audits and hypothesis generation: it can show which controlled state variables are linearly accessible from a model-plus-adapter system under a shared readout. Do not use this first snapshot for architecture rankings, deployment claims, or claims about transfer to operational tasks. The calibration table covers metric floor, raw, random, FFT, statistical, and oracle feature rows at five lengths, but stronger model-quality claims still need supervised models trained from scratch, target-specific closed-form estimators, fixed-length ablations of pretrained encoders, and held-out process/view families. The probe protocol measures linear accessibility from mean-pooled frozen representations, so negative results do not rule out token-level probes, learned pooling, nonlinear readouts, manifold analyses, model-specific heads, or fine-tuning. Aionoscope is therefore most decisive in one direction. A probe \emph{success} certifies that a state variable is linearly accessible from the frozen representation under a cheap shared readout; a probe \emph{failure} indicates only inaccessibility under this pooled linear readout, not absent information. Practitioners should treat failures as hypotheses for richer-readout or fine-tuning follow-up rather than grounds to discard a layer or model, and should weight positive findings more strongly than negative ones. Under this calibrated but native-length pooled-linear-probe design, the evidence supports Aionoscope as a public validation-seed pilot diagnostic snapshot; it does not establish length-invariant orderings, protection against benchmark overfitting, absence of unrecovered information, or real-task transfer. We also do not report runtime, latency, throughput, or memory claims from the current benchmark runs.

The PDF, browser, and source artifacts expose this first public validation-seed snapshot and reproduction material for the generator, adapters, checkpoint IDs, manifests, and scripts. These seeds now become public reference streams; future benchmark claims should use hidden or locked evaluation streams. We will maintain public releases and should archive camera-ready artifacts with durable identifiers, result hashes, and checkpoint manifests. The streams are synthetic, contain no private or human-subject data, and do not support safety-critical deployment claims; checkpoints retain their original licenses and intended-use constraints.

\section{Conclusion}

The current results support Aionoscope as a hypothesis-generating diagnostic scaffold, not as a validated predictor of real-task performance. Under one deterministic primitive-mixture generator and one native-length pooled linear-probe protocol, Aionoscope gives a diagnostic view of model-plus-adapter accessibility rather than a length-controlled ranking of encoders. The main lesson is that coarse component recovery is not the same as latent-state accessibility: a representation can expose that a signal contains a periodic, trending, noisy, or event-like component while failing to expose the dense timing, phase, amplitude, frequency, or regime variables needed for debugging. Target-family, mixture-complexity, and layer-wise views make this failure mode visible in ways that final-layer-only evaluations can miss. These results motivate broader Aionoscope instances with richer readout studies, hidden evaluation streams, held-out process/view families, and real-data validation.

\bibliographystyle{ACM-Reference-Format}
\bibliography{references}

\end{document}